\documentclass[journal]{IEEEtran}

\usepackage{times}
\usepackage{epsfig}
\usepackage{graphicx}
\usepackage{amsmath}
\usepackage{amssymb}
\usepackage{amsthm}
\usepackage{setspace}
\usepackage{algorithm}
\usepackage{algorithmic}
\usepackage{color}
\usepackage{hyperref}
\usepackage{dsfont}
\usepackage{cases}
\usepackage{booktabs}

\theoremstyle{plain}

\begin{document}

\title{A Regularization Approach for Instance-based Superset Label Learning}
%
%
%

\author{Chen~Gong,~\IEEEmembership{}
        Tongliang~Liu,~\IEEEmembership{}
        Yuanyan~Tang,~\IEEEmembership{Fellow,~IEEE,}
        Jian~Yang,~\IEEEmembership{ }
        Jie~Yang,~\IEEEmembership{  }
        Dacheng~Tao,~\IEEEmembership{Fellow,~IEEE,}
\thanks{This research is supported by NSFC of China (No. 61602246, 61572315, 91420201,
61472187, 61502235, 61233011 and 61373063), 973 Plan of China (No. 2014CB349303 and 2015CB856004), Program for Changjiang Scholars, and Australian Research Council Projects (No. DP-140102164, FT-130101457 and LP-150100671).}
\thanks{Chen Gong is with the School of Computer Science and Engineering, Nanjing University of Science and Technology, and also with the Institute of Image Processing and Pattern Recognition, Shanghai Jiao Tong University. (e-mail: chen.gong@njust.edu.cn).}
\thanks{Tongliang Liu is with the School of Software, Faculty of Engineering and Information Technology, University of Technology Sydney, Ultimo, NSW 2007, Australia (e-mail: tliang.liu@gmail.com).}
\thanks{Yuanyan Tang is with the Faculty of Science and Technology, University of
Macau, Macau 999078, China, and also with the College of Computer Science,
Chongqing University, Chongqing 400000, China (e-mail: yytang@umac.mo;
yytang@cqu.edu.cn).}
\thanks{Jian Yang is with School of Computer Science and Engineering, Nanjing University of Science and Technology, Nanjing, China, 210094 (e-mail: csjyang@njust.edu.cn).}
\thanks{Jie Yang is with the Institute of Image Processing and Pattern Recognition, Shanghai Jiao Tong University, Shanghai, China, 200240 (e-mail: jieyang@sjtu.edu.cn).}
\thanks{Dacheng Tao is with the School of Information Technologies and the Faculty of Engineering and Information Technologies, University of Sydney, J12/318 Cleveland St, Darlington NSW 2008, Australia (e-mail: dacheng.tao@sydney.edu.au).}
\thanks{\textcopyright20XX IEEE. Personal use of this material is permitted. Permission from IEEE must be obtained for all other uses, in any current or future media, including reprinting/republishing this material for advertising or promotional purposes, creating new collective works, for resale or redistribution to servers or lists, or reuse of any copyrighted component of this work in other works.}
}


\maketitle

\begin{abstract}
  Different from the traditional supervised learning in which each training example has only one explicit label, \textbf{S}uperset \textbf{L}abel \textbf{L}earning (SLL) refers to the problem that a training example can be associated with a set of candidate labels, and only one of them is correct. Existing SLL methods are either regularization-based or instance-based, and the latter of which has achieved state-of-the-art performance. This is because the latest instance-based methods contain an explicit disambiguation operation that accurately picks up the groundtruth label of each training example from its ambiguous candidate labels. However, such disambiguation operation does not fully consider the mutually exclusive relationship among different candidate labels, so the disambiguated labels are usually generated in a non-discriminative way, which is unfavorable for the instance-based methods to obtain satisfactory performance. To address this defect, we develop a novel \textbf{Reg}ularization approach for \textbf{I}nstance-based \textbf{S}uperset \textbf{L}abel (RegISL) learning so that our instance-based method also inherits the good discriminative ability possessed by the regularization scheme. Specifically, we employ a graph to represent the training set, and require the examples that are adjacent on the graph to obtain similar labels. More importantly, a discrimination term is proposed to enlarge the gap of values between possible labels and unlikely labels for every training example. As a result, the intrinsic constraints among different candidate labels are deployed, and the disambiguated labels generated by RegISL are more discriminative and accurate than those output by existing instance-based algorithms. 
  The experimental results on various tasks convincingly demonstrate the superiority of our RegISL to other typical SLL methods in terms of both training accuracy and test accuracy.
\end{abstract}


\IEEEpeerreviewmaketitle

\section{Introduction}
\label{sec:intro}
In \textbf{S}uperset \textbf{L}abel \textbf{L}earning (SLL), one training example can be ambiguously labeled with multiple candidate labels, among which only one is correct. This is different from the conventional supervised classification which works on the training examples with each of them only has one explicit label. \par

\begin{figure*}
  \centering
  \includegraphics[width=\linewidth]{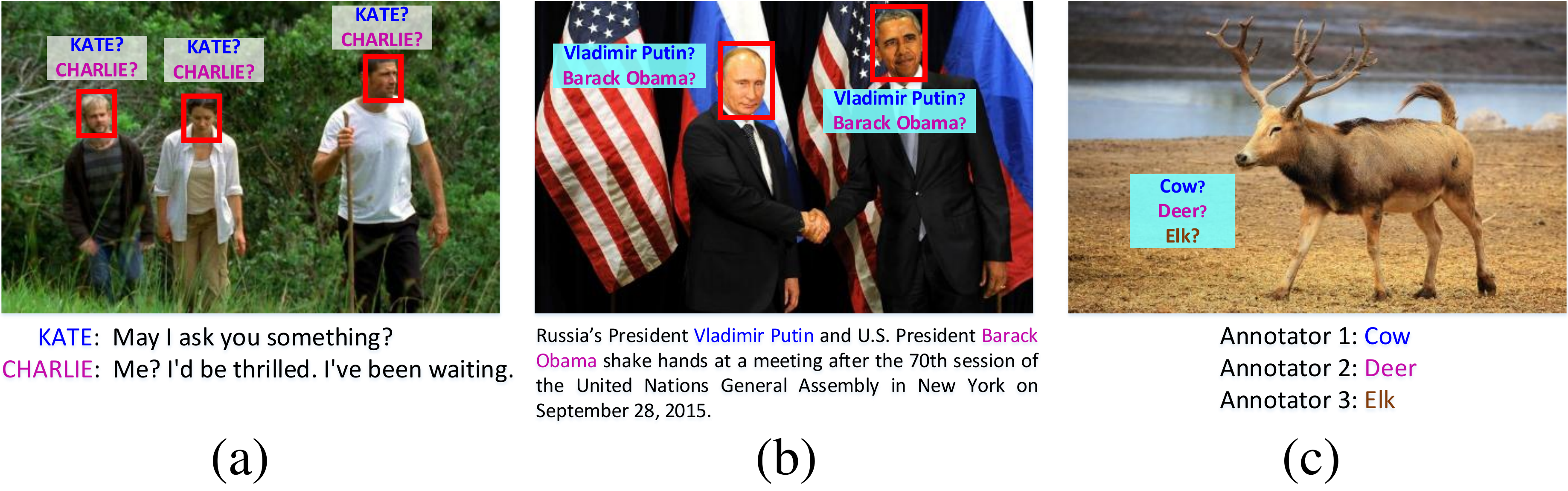}\\
  \caption{Some example applications of superset label learning. (a) is a screenshot of ``Lost'' TV serial (Season 1), in which three characters' faces are detected. From the scripts provided below, we can infer that both Kate and Charlie appear in this screenshot. However, it still remains unclear that which face corresponds to Kate and which face belongs to Charlie. (b) shows a news image and its description from the news website ``http://fox17online.com/''. From the textual description we know that these two people are Vladimir Putin and Barack Obama. However, which face corresponds to Vladimir Putin or Barack Obama is not clearly indicated. (c) shows an image of elk, which is an animal very similar to both cow and deer. In the application of crowdsourcing, the involved annotators may have different levels of expertise, so different labels are possibly provided by the different annotators, which can be either correct or incorrect.}\label{fig:PLLIlllustration}
\end{figure*}

SLL has a variety of applications. For example, an episode of a video or TV serial may contain several characters chatting with each other, and their faces may appear simultaneously in a screenshot. We also have access to the scripts and dialogues indicating the characters' names. However, these information only reveals who are in the given screenshot, but does not build the specific one-to-one correspondence between the characters' faces and the appeared names. Therefore, each face in the screenshot is ambiguously named, and our target is to determine the groundtruth name of each face in the screen shot (see Fig.~\ref{fig:PLLIlllustration}(a)). Another similar application is that in a photograph collection such as newsletters or family album, each photo may be annotated with a description indicating who are in this photo. However, the detailed identity of each person in the photo is not specified, so matching the persons with their real names is useful
(see Fig.~\ref{fig:PLLIlllustration}(b)). SLL problem also arises in crowdsourcing, in which each example (image or text) is probably assigned multiple labels by different annotators. Nevertheless, some of the labels may be incorrect or biased because of the difference among various annotators in terms of expertise or cultural background, so it is necessary to find the most suitable label of every example resided in the candidate labels (see Fig.~\ref{fig:PLLIlllustration}(c)). In above applications, manually labeling the groundtruth label of each example will incur unaffordable monetary or time cost, so SLL can be an ideal tool for tackling such problems with ambiguously labeled examples. \par

Superset label learning \cite{liu2012conditional} is also known as ``partial label learning'' \cite{cour2011learning,zhangsolving2015,yuACML15} and ``ambiguously label learning'' \cite{hullermeier2006learning,chen2014ambiguously}. For the consistency of our presentation, we will use the term ``superset label learning'' throughout this paper. Superset label learning is formally defined as follows. Suppose we have $n$ training examples $\mathcal{X}=\{\mathbf{x}_1,\mathbf{x}_2,\cdots,\mathbf{x}_n\}\in\mathbb{R}^{d}$ with dimensionality $d$, and their candidate labels are recorded by $n$ label sets $\mathcal{S}_1,\mathcal{S}_2,\cdots,\mathcal{S}_n$, respectively. Therefore, the entire candidate label space consisted of $c$ possible class labels has the size $2^c$. Besides, we assume that the groundtruth labels of these $n$ training examples are $y_1,y_2,\cdots,y_n$ with $y_i\in \mathcal{S}_i$ ($i=1,2,\cdots,n$), whereas they are unknown to the learning algorithms. Therefore, given the output label set denoted by $\mathcal{Y}=\left\{1,2,\cdots,c\right\}$, the target of a SLL algorithm is to build a classifier $f$ based on $\mathcal{X}$ so that it can accurately predict the single unambiguous label $y_t\in\mathcal{Y}$ of an unseen test example $\mathbf{x}_t$.\par

\subsection{Related Work}
\label{sec:related_work}

To the best of our knowledge, the concept of SLL was firstly proposed by Grandvalet \cite{grandvallet2002logistic}, who elegantly adapts the traditional logistic regression to superset label cases. After that, there are mainly two threads for tackling the SLL problem: regularization-based models and instance-based models.

\subsubsection{Regularization-based Models}

Regularization-based models try to achieve maximum margin effect by developing various loss functions. For example, Jin et al. \cite{jin2002learning} firstly assume that every element in the candidate set $\mathcal{S}_i$ ($i=1,2,\cdots,n$) has equal probability to be the correct label, and designs a ``naive'' superset label loss. Next, considering that it is inappropriate to treat all the candidate labels equally, they further propose to disambiguate the candidate labels, \emph{i.e.} directly discovering each example's groundtruth label from its multiple candidate labels, so that a discriminative loglinear model can be built. 
Besides, Cour et al. \cite{cour2011learning,cour2009learning} hold that the above naive loss is loose compared to the real superset label 0-1 loss $\mathcal{L}_{01}\left(f(\mathbf{x}_i),\mathcal{S}_i\right)=\mathds{1}\left[f(\mathbf{x}_i)\notin \mathcal{S}_i\right]$\footnote{The operation ``$\mathds{1}\left[\cdot\right]$'' returns 1 if the argument within the bracket holds true, and 0 otherwise.}, so they propose another novel surrogate loss that is a tighter approximation to the real 0-1 loss than the naive loss. To be specific, this loss function is formulated as $\mathcal{L}\left(f(\mathbf{x}_i),\mathcal{S}_i\right)=\Psi\left[\frac{1}{|\mathcal{S}_i|}\sum_{j\in \mathcal{S}_i}f_j(\mathbf{x}_i)\right]+\sum_{j\notin\mathcal{S}_i}\Psi\left[-f_j(\mathbf{x}_i)\right]$ where $\Psi\left[\cdot\right]$ can be hinge, exponential or logistic loss. Here the first term computes the mean value of the scores $f_j(\mathbf{x}_i)$ of the labels in $\mathcal{S}_i$. However, this averaging strategy has a critical shortcoming that its effectiveness can be largely decreased by the false positive label(s) $\mathcal{S}_i- y_i$ in the candidate label set $\mathcal{S}_i$. As a result, the training process will be dominated by these false positive labels and the final model output can be biased. 
Therefore, Nguyen et al. \cite{nguyen2008classification} develop the superset label hinge loss that maximizes the margin between the maximum model output among candidate labels and that among the remaining non-candidate labels, namely $\mathcal{L}\left(f(\mathbf{x}_i),\mathcal{S}_i\right)=\max\left(0,1-\left[\max\limits_{y_i\in\mathcal{S}_i}f(\mathbf{x}_i,y_i;\omega)
-\max\limits_{y^{\prime}_i\notin\mathcal{S}_i}f(\mathbf{x}_i,y^{\prime}_i;\omega)\right]\right)$ where $\omega$ is the model parameter. Differently, H{\"u}llermeier et al. \cite{H2015Superset} propose a generalized loss with its expression $\mathcal{L}(f(\mathbf{x}_i),\mathcal{S}_i)=\min_{y_i\in \mathcal{S}_i }\Psi[y_i,f(\mathbf{x}_i)]$, where $\Psi[\cdot]$ represents the logistic loss. However, above two formulations do not discriminate the groundtruth label $y_i$ from other candidate labels. Therefore, Yu et al. \cite{yuACML15} devise a new SLL maximum margin formulation based on Support Vector Machines (SVM) which directly maximizes the margin between the groundtruth label and all other labels. The corresponding loss function is $\mathcal{L}\left(f(\mathbf{x}_i),\mathcal{S}_i\right)=f(\mathbf{x}_i,y_i;\omega)-\max\limits_{y^{\prime}_{i}\neq y_i}f(\mathbf{x}_i,y^{\prime}_i;\omega)$. Different from above methods that only assume that one example is associated with a set of candidate labels, Luo et al. \cite{luo2010learning} consider a generalized setting in which each training example is a bag containing multiple instances and is associated with a set of candidate label vectors. Each label vector encodes the possible labels for the instances in the bag, and only one of them is fully correct.\par

For the theoretical aspect, Cid-Sueiro \cite{cid2012proper} studies the general necessary and sufficient condition for designing a SLL loss function, and provide a detailed procedure to construct a proper SLL loss under practical situations. Cid-Sueiro et al. \cite{Cid2014Consistency} also reveal that the consistency of loss functions depends on the mixing matrix, which refers to the transition matrix relating the candidate labels and the groundtruth label. More generally, Liu et al. \cite{liu2014learnability} discuss the learnability of regularization-based SLL approaches, and reveal that the key to achieving learnability is that the expected classification error of any hypothesis in the space can be bounded by the superset label 0-1 loss averaged over the entire training set.\par

Other representative regularization-based SLL algorithms include \cite{chen2014ambiguously,shrivastava2012learning,zhang2014disambiguation} that utilize coding theory, \cite{liu2012conditional} that employs the conditional multinomial mixture model, and \cite{zeng2013learning} that leverages the low-rank assumption \cite{Xu2016Local,Xu2015MultiTIP} to capture the example-label correspondences.

\subsubsection{Instance-based Models}

Instance-based models usually construct a nonparametric classifier on the training set, and the candidate label set of a training example can be either disambiguated or kept ambiguous as it originally presents. H{\"u}llermeier et al. \cite{hullermeier2006learning} propose a series of nonparametric models such as superset label $K$-nearest neighborhood classifier and decision tree. The models in \cite{hullermeier2006learning} do not have a disambiguation operation and directly use the ambiguous label sets for training and testing. Differently, Zhang et al. \cite{zhangsolving2015} proposes an iterative label propagation scheme to disambiguate the candidate labels of training examples. 
Furthermore, considering that the disambiguation process in current methods simply focuses on manipulating the label space, Zhang et al. \cite{Zhang2016Partial} advocate making full use of the manifold information \cite{Gong2015Deformed} embedded in the feature space, and propose a feature-aware disambiguation.


\subsection{Our Motivation}
\label{sec:motivation}
Although the method proposed in \cite{zhangsolving2015} generally obtains the best performance among all existing SLL algorithms, it still suffers from several drawbacks. Firstly, as an instance-based method, it falls short of discovering the mutually exclusive relationship among different candidate labels, and does not take specific measures to highlight the potential groundtruth label during the disambiguation process. Secondly, as an iterative algorithm, the convergence property of the propagation sequence is only empirically illustrated and does not have a theoretical guarantee.\par

To address above two shortcomings, we propose a \textbf{Reg}ularization
approach for \textbf{I}nstance-based \textbf{S}uperset \textbf{L}abel learning, and term it as ``RegISL''. The advantages of our RegISL are two folds: Firstly, to make the disambiguated labels discriminative, we design a proper discrimination regularizer along with the related constraints to increase the gap of scores between possible candidate labels and unlikely candidate labels. As a result, the potential groundtruth labels will become prominent, whereas the unlikely labels will be suppressed. Secondly, to avoid the convergence problem of iterative algorithm like \cite{zhangsolving2015}, we solve the designed optimization problem via the Augmented Lagrangian Multiplier (ALM) method \cite{Xu2015MultiPAMI,Gong2016TLLT} which will always finds a stationary solution. Besides, due to the nonconvexity of the augmented Lagrangian objective function, we show that it can be decomposed as the difference of two convex components and then minimized by the ConCave Convex Procedure (CCCP) \cite{yuille2003concave}.\par

We empirically test our RegISL and other representative SLL methodologies \cite{liu2012conditional,cour2011learning,zhangsolving2015,yuACML15,hullermeier2006learning,zhang2014disambiguation} on various practical applications such as character-name association in TV show, ambiguous image classification, automatic face naming in news images, and bird sound classification. The experimental results suggest that in most cases the proposed RegISL is able to outperform other competing baselines in terms of both training accuracy and test accuracy.

\section{Model Description}
\label{sec:model}
\begin{figure}
  \centering
  \includegraphics[width=\linewidth]{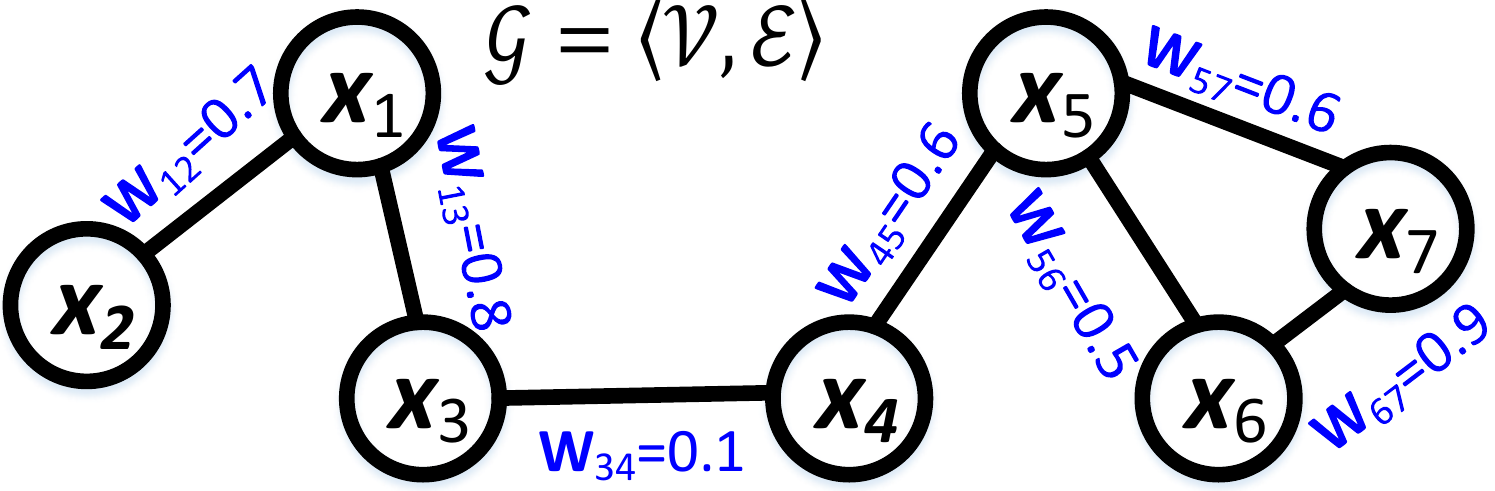}\\
  \caption{The illustration of graph $\mathcal{G}=\langle\mathcal{V},\mathcal{E}\rangle$, where in this example the seven circles represent the node set $\mathcal{V}=\left\{\mathbf{x}_1,\mathbf{x}_2,\cdots,\mathbf{x}_7\right\}$, and the lines connecting them constitute the edge set $\mathcal{E}=\left\{\mathbf{W}_{12},\mathbf{W}_{13},\cdots,\mathbf{W}_{67}\right\}$. The edge weights $\mathbf{W}_{ik}$ ($i,k=1,2,\cdots,7$) are indicated besides the edges in blue color.}\label{fig:GraphIllustration}
\end{figure}

This section introduces our nonparametric instance-based method RegISL. In the training stage (Section~\ref{sec:training}), a graph $\mathcal{G}=\langle\mathcal{V},\mathcal{E}\rangle$ is established on the training set to capture the relationship between pairs of training examples, where $\mathcal{V}$ is the node set representing all $n$ training examples and $\mathcal{E}$ is the edge set encoding the similarities between these nodes (see Fig.~\ref{fig:GraphIllustration}). In this work, two examples $\textbf{x}_i$ and $\textbf{x}_k$ are linked by an edge in $\mathcal{G}$ if one of them belongs to the $K$ nearest neighbors of the other one, and the edge weight (\emph{i.e.} the similarity between $\textbf{x}_i$ and $\textbf{x}_k$) is computed by the Gaussian kernel function \cite{xiao2015parameterGaussian,Gong2016MultiTLLT}
\begin{equation}\label{eq_Gaussian}
  {{\mathbf{W}}_{ik}}=\exp \left( -\frac{\left\| {{\mathbf{x}}_{i}}-{{\mathbf{x}}_{k}} \right\|^{2}}{2{{\theta }^{2}}} \right),
\end{equation}
where $\theta$ denotes the kernel width. In contrast, $\mathbf{W}_{ik}$ is set to 0 if there is no edge between $\mathbf{x}_i$ and $\mathbf{x}_k$. After that, a regularized objective function is built on $\mathcal{G}$, which is able to disambiguate the candidate labels and discover the unique real label of every training example. In the test stage (Section~\ref{sec:test}), the test example $\mathbf{x}_t$ is assigned label $y_t$ ($y_t$ takes a value from $1,2,\cdots,c$ with $c$ being the total number of classes) based on the disambiguated labels of its $K$ nearest neighbors in the training set.

\subsection{Training Stage}
\label{sec:training}
For our instance-based RegISL, the main target of training stage is to pick up the real label $y_i$ of each training example $\mathbf{x}_i$ from its candidate label set $\mathcal{S}_i$. The established graph $\mathcal{G}$ can be quantified by the adjacency matrix $\mathbf{W}$ where its $(i,k)$-th element is $\textbf{W}_{ik}$ if $i\neq k$ and 0 otherwise \cite{zhu2003semi,wang2016semi}.\par

Similar to \cite{zhangsolving2015}, the candidate labels of a training example $\mathbf{x}_i$ ($i$ takes a value from $1,2,\cdots, n$) is represented by a $c$-dimensional label vector $\mathbf{Y}_i$, which is
\begin{equation}\label{eq1}
\mathbf{Y}_{ij}:=\left\{ \begin{split}
  & 1/|\mathcal{S}_i|, \quad  \mathbf{x}_i \text{ has the candidate label } j \\
  & 0, \qquad \quad \text{otherwise} \\
\end{split} \right.,
\end{equation}
where $|\mathcal{S}_i|$ denotes the size of set $\mathcal{S}_i$. Note that the sum of all the elements in every $\mathbf{Y}_i$ is 1 according to the definition of \eqref{eq1}. Furthermore, we use the vectors $\mathbf{F}_1,\mathbf{F}_2,\cdots,\mathbf{F}_n\in \mathbb{R}^{1\times c}$ to record the obtained labels of training examples $\mathbf{x}_1,\mathbf{x}_2,\cdots,\mathbf{x}_n$, respectively, in which $\mathbf{F}_{ij}$ can be understood as the probability of $\mathbf{x}_{i}$ belonging to the class $j$, then our regularization model for RegISL can be expressed as
\begin{equation}\label{eq_addmodel}
\begin{split}
  & \underset{{{\mathbf{F}}_{1}},\cdots ,{{\mathbf{F}}_{n}}}{\mathop{\min }}\ \frac{1}{2}\sum\limits_{i=1}^{n}{\sum\limits_{k=1}^{n}{{{\mathbf{W}}_{ik}}\left\| {{\mathbf{F}}_{i}}\!-\!{{\mathbf{F}}_{k}} \right\|_{2}^{2}}}+\alpha \sum\limits_{i=1}^{n}{\sum\limits_{j\in {{\Omega }_{i}}}\!{{{({{\mathbf{F}}_{ij}}\!-\!{{\mathbf{Y}}_{ij}})}^{2}}}}\\
  &\qquad \qquad -\beta \sum\limits_{i=1}^{n}{\left\| {{\mathbf{F}}_{i}} \right\|_{2}^{2}} \\
 & \quad s.t.\ \sum\limits_{j=1}^{c}{{{\mathbf{F}}_{ij}}=1},\ {{\mathbf{F}}_{ij}}\ge 0,\ \ \forall \ i=1,2,\cdots ,n \\
\end{split}\!\!.
\end{equation}
In Eq.~\eqref{eq_addmodel}, the set ${\Omega }_{i}$ includes the subscripts of zero elements in $\mathbf{Y}_i$, ``$\left\|\cdot\right\|_2$'' computes the $l_2$ norm of the vector, and $\alpha$ and $\beta$ are nonnegative trade-off parameters controlling the relative weights of the three terms in the objective function.\par

The first term in the objective function of Eq.~\eqref{eq_addmodel} is called \emph{smoothness term}, which requires the two examples connected by a strong edge (\emph{i.e.} the edge weight is large) in $\mathcal{G}$ to obtain similar labels \cite{zhu2003semi,gong2014fick,pei2015manifold}, so minimizing this smoothness term will force $\mathbf{F}_i$ to get close to $\mathbf{F}_k$ if $\mathbf{W}_{ik}$ is large. The second term is called \emph{fidelity term}, which suggests that if $\mathbf{x}_i$'s candidate label set $\mathcal{S}_i$ does not contain the label $j$ (\emph{i.e.} $\mathbf{Y}_{ij}=0$), then the $j$-th element in the finally obtained label vector $\mathbf{F}_{i}$ should also be zero. Although there are many other ways to character the difference between $\mathbf{F}_{ij}$ and $\mathbf{Y}_{ij}$, here we simply adopt the quadratic form as it is perhaps the simplest way to compare $\mathbf{F}_{ij}$ and $\mathbf{Y}_{ij}$. This form has also been widely used by many semi-supervised learning methodologies such as \cite{gong2014fick,Zhou03learningwith,wang2009linear}. The third \emph{discrimination term} along with the \emph{normalization constraint} $\sum\nolimits_{j=1}^{c}{{{\mathbf{F}}_{ij}}=1}$ and \emph{nonnegative constraint} ${{\mathbf{F}}_{ij}}\ge 0$, critically makes the obtained $\mathbf{F}_i$ to be discriminative. That is to say, by requiring the elements in $\mathbf{F}_i$ nonnegative and summing up to 1, minimizing $-\left\|\mathbf{F}_i\right\|_2^2$ (\emph{i.e.} maximizing $\left\|\mathbf{F}_i\right\|_2^2$) will widen the gap of values between possible labels and unlikely labels of $\mathbf{x}_i$, and thus yielding discriminative and confident label vector $\mathbf{F}_i$. The detailed reasons are explained as follows.\par

\begin{figure}
  \centering
  \includegraphics[width=0.95\linewidth]{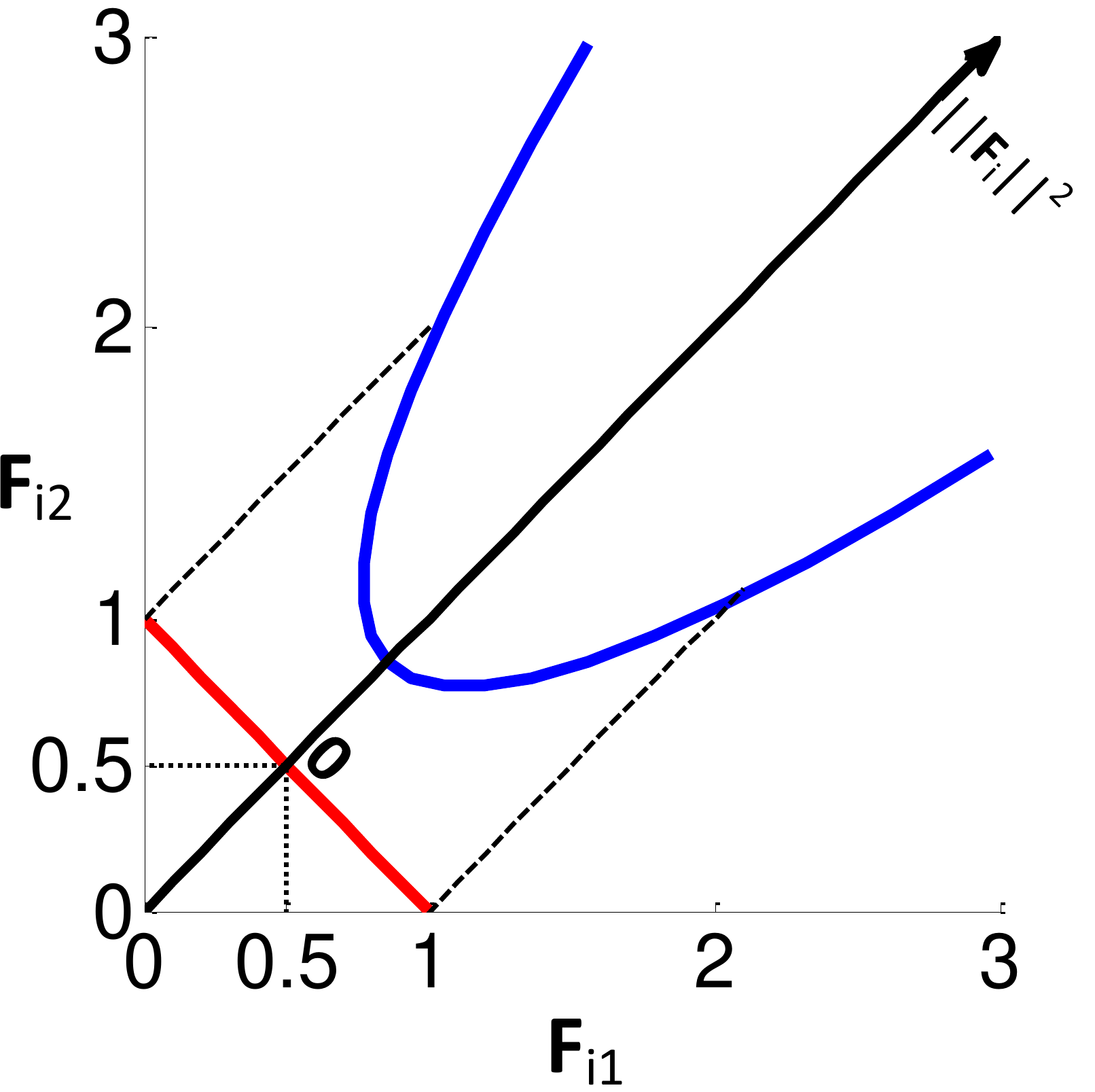}\\
  \caption{The motivation of our introduced discrimination term along with the nonnegative and normalization constraints. Suppose the example $\mathbf{x}_i$'s label vector is $\mathbf{F}_{i}=[\mathbf{F}_{i1},\mathbf{F}_{i2}]$, then the valid outputs of $\mathbf{F}_i$ satisfying the constrains in Eq.~\eqref{eq_addmodel} are on the red line $\mathbf{F}_{i1}+\mathbf{F}_{i2}=1$ ($\mathbf{F}_{i1}\geq0$, $\mathbf{F}_{i2}\geq0$). Taking this red curve as x-axis and $(0.5,0.5)$ as original point, the value of $\left\|\mathbf{F}_i\right\|_{2}^{2}$ with varying $\mathbf{F}_{i1}$ and $\mathbf{F}_{i2}$ is recorded by the blue curve. We observe that the smallest $\left\|\mathbf{F}_i\right\|_{2}^{2}$ corresponds to the most ambiguous label vector $[0.5, 0.5]$, while $\left\|\mathbf{F}_i\right\|_{2}^{2}$ becomes large when $\mathbf{F}_{i}=[\mathbf{F}_{i1},\mathbf{F}_{i2}]$ gets close to the discriminative results $[1,0]$ and $[0,1]$.}\label{fig:DiscriminationIllustration}
\end{figure}

Suppose that we are dealing with a binary classification problem (\emph{i.e.} $c=2$), and the label vector of example $\mathbf{x}_i$ is $\mathbf{F}_{i}=[\mathbf{F}_{i1},\mathbf{F}_{i2}]$. If $\mathbf{x}_i$ is initially associated with the ambiguous candidate labels 1 and 2 (\emph{i.e.} $\mathbf{Y}_{ij}=[0.5, 0.5]$), we hope that the finally obtained $\mathbf{F}_i$ can approach to $[1, 0]$ or $[0, 1]$, which confidently implies that $\mathbf{x}_i$ belongs to the first or second class. In contrast, the output close to $\mathbf{F}_i=[0.5, 0.5]$ is not encouraged because such $\mathbf{F}_i$ does not convey any information for deciding $\mathbf{x}_i$'s real label. To this end, we impose the nonnegative and normalization constrains on $\mathbf{F}_{i}$ as in Eq.~\eqref{eq_addmodel}, then its elements $\mathbf{F}_{i1}$ and $\mathbf{F}_{i2}$ will only select the values along the red line in Fig.~\ref{fig:DiscriminationIllustration}. Furthermore, we take the red line as x-axis and plot the squared $l_2$ norm of $\mathbf{F}_i$ under different $\mathbf{F}_{i1}$ and $\mathbf{F}_{i2}$ (see the blue curve). It can be clearly observed that $\left\|\mathbf{F}_i\right\|_{2}^{2}$ hits the lowest value when both $\mathbf{F}_{i1}$ and $\mathbf{F}_{i2}$ are equal to 0.5, and $\left\|\mathbf{F}_i\right\|_{2}^{2}$ gradually increases when $[\mathbf{F}_{i1},\mathbf{F}_{i2}]$ approaches to $[0,1]$ or $[1,0]$. Therefore, the label vector $\mathbf{F}_i$ with large norm is encouraged by the discrimination term in Eq.~\eqref{eq_addmodel}, so that the obtained $\mathbf{F}_i$ prefers definite results $[0,1]$ or $[1,0]$ and meanwhile avoids the ambiguous outputs that are close to $[0.5, 0.5]$.\par

For ease of optimizing Eq.~\eqref{eq_addmodel}, we may reformulate it into a compact formation. Based on $\mathcal{G}$'s adjacency matrix $\mathbf{W}$, we further define a diagonal degree matrix $\mathbf{D}$ with the $i$-th diagonal element representing $\mathbf{x}_i$'s degree computed by ${{\mathbf{D}}_{ii}}=\sum\nolimits_{j=1}^{n}{{{\mathbf{W}}_{ij}}}$. Therefore, a positive semi-definite graph Laplacian matrix can be calculated as $\mathbf{L}=\mathbf{D}-\mathbf{W}$. Besides, we stack the row vectors $\mathbf{Y}_1,\mathbf{Y}_2,\cdots,\mathbf{Y}_n$ as $\mathbf{Y}={{\left( \mathbf{Y}_{1}^{\top},\mathbf{Y}_{2}^{\top},\cdots ,\mathbf{Y}_{n}^{\top} \right)}^{\top}}$ to establish a $n\times c$ candidate label matrix $\mathbf{Y}$. Similarly, the label matrix $\mathbf{F}$ to be optimized is established by $\mathbf{F}={{\left( \mathbf{F}_{1}^{\top},\mathbf{F}_{2}^{\top},\cdots ,\mathbf{F}_{n}^{\top} \right)}^{\top}}$. Furthermore, by defining $\mathbf{1}_c$, $\mathbf{1}_n$ and $\mathbf{O}_{n\times c}$ as the $c$-dimensional all-one vector, $n$-dimensional all-one vector, and $n\times c$-dimensional all-zero matrix, respectively, Eq.~\eqref{eq_addmodel} can be rewritten as
\begin{equation}\label{eq2}
\begin{split}
  & \min_{\mathbf{F}} \quad tr(\mathbf{F}^{\top}\mathbf{LF})+\alpha\left\|\mathbf{H}\odot(\mathbf{F}-\mathbf{Y}) \right\|_\mathrm{F}^2-\beta\left\|\mathbf{F}\right\|_\mathrm{F}^2 \\
  & \ s.t. \quad \mathbf{F1}_c=\mathbf{1}_n,\quad \mathbf{F}\geq\mathbf{O}_{n\times c}\\
\end{split}.
\end{equation}
In Eq.~\eqref{eq2}, ``$\left\|\cdot\right\|_\mathrm{F}$'' computes the Frobenius norm of corresponding matrix, and ``$\odot$'' refers to the elementwise product. $\mathbf{H}$ is a $\{0,1\}$-binary matrix with the element $\mathbf{H}_{ij}=1$ if $\mathbf{Y}_{ij}=0$ and 0 otherwise.\par
Since Eq.~\eqref{eq2} is a constrained optimization problem, we may use the method of Augmented Lagrangian Multiplier (ALM) to find its solution. Compared to the traditional Lagrangian method, ALM adds an additional quadratic penalty function to the objective, which leads to faster convergence rate and lower computational cost \cite{bertsekas2014constrained}. Therefore, by introducing the multipliers $\mathbf{\Lambda}_1$ and $\mathbf{\Lambda}_2$ to deal with the nonnegative constraint and normalization constraint, respectively, the augmented Lagrangian function is expressed as
\begin{equation}\label{eq3}
\begin{split}
  &J(\mathbf{F},\mathbf{\Lambda}_1,\mathbf{\Lambda}_2,\sigma)=tr(\mathbf{F}^{\top}\mathbf{LF})+\alpha\left\|\mathbf{H}\odot(\mathbf{F}-\mathbf{Y}) \right\|_\mathrm{F}^2\\
  &\quad-\beta\left\|\mathbf{F}\right\|_\mathrm{F}^2
  +\frac{1}{2\sigma}tr(\mathbf{M}^{\top}\mathbf{M}-\mathbf{\Lambda}_1^{\top}\mathbf{\Lambda}_1)-\mathbf{\Lambda}_2^{\top}(\mathbf{F1}_c-\mathbf{1}_n)\\
  &\quad+\frac{\sigma}{2}\left\|\mathbf{F1}_c-\mathbf{1}_n\right\|_2^2\\
\end{split},
\end{equation}
where $\mathbf{M}= \max \left\{\mathbf{O}_{n\times c},\mathbf{\Lambda}_1-\sigma\mathbf{F}\right\}$ is an auxiliary variable that enforces the obtained optimal $\mathbf{F}$ (\emph{i.e.} $\mathbf{F}^{\star}$) to be nonnegative. The operation ``$\max(\mathbf{A},\mathbf{B})$'' returns a matrix with its $(i,j)$-th element being the largest element between $\mathbf{A}_{ij}$ and $\mathbf{B}_{ij}$. The variable $\sigma>0$ is the penalty coefficient.\par
Based on Eq.~\eqref{eq3}, the optimal solution of Eq.~\eqref{eq2} can be obtained by alternately updating $\mathbf{F}$, $\mathbf{\Lambda}_{1}$, $\mathbf{\Lambda}_{2}$ and $\sigma$, among which $\mathbf{\Lambda}_{1}$, $\mathbf{\Lambda}_{2}$ and $\sigma$ can be easily updated via the conventional rules of ALM, namely:
\begin{numcases}
  \mathbf{\Lambda}_{1}:=\max\left\{\mathbf{O}_{n\times c},\mathbf{\Lambda}_{1}-\sigma\mathbf{F}\right\} \label{eq4}\\
  \mathbf{\Lambda}_{2}:=\mathbf{\Lambda}_{2}-\sigma(\mathbf{F}\cdot\mathbf{1}_c-\mathbf{1}_n) \label{eq5}\\
  \sigma:=\min\left\{\rho\sigma,10^8\right\} \label{eq6}
\end{numcases}
In \eqref{eq6}, the operation ``$\min$'' selects the smallest value in the bracket, and $\rho=1.1$ is the parameter that makes $\sigma$ gradually increase in each loop so that the normalization constraint can be finally satisfied.\par
However, the updating of $\mathbf{F}$ is difficult because Eq.~\eqref{eq3} regarding $\mathbf{F}$ is nonconvex due to the nonpositive term $-\beta\left\|\mathbf{F}\right\|_{\mathrm{F}}^2$. Therefore, we use the method of ConCave Convex Procedure (CCCP) proposed by Yuille et al. \cite{yuille2003concave} to update $\mathbf{F}$. CCCP can be regarded as a majorization-minimization algorithm \cite{sriperumbudur2011majorization} that solves the original nonconvex problem as a sequence of convex programming. Specifically, the main idea of CCCP is to decompose the nonconvex objective function $J(\mathbf{F})$ as the difference of two convex functions $J_1(\mathbf{F})$ and $J_2(\mathbf{F})$, namely $J(\mathbf{F})=J_1(\mathbf{F})-J_2(\mathbf{F})$; and in each iteration $J_2(\mathbf{F})$ is replaced by its first order Taylor approximation $\tilde{J}_2(\mathbf{F})$, and the original objective function $J(\mathbf{F})$ is then approximated by the convex $J(\mathbf{F})=J_1(\mathbf{F})-\tilde{J}_2(\mathbf{F})$. Theoretical analyses suggest that CCCP is always able to converge to a local minima \cite{lanckriet2009convergence}. In our case, we choose the two convex functions $J_1(\mathbf{F})$ and $J_2(\mathbf{F})$ as
\begin{equation}\nonumber
  \left\{
  \begin{split}
  J_1(\mathbf{F})=&tr(\mathbf{F}^{\top}\mathbf{LF})+\alpha\left\|\mathbf{H}\odot(\mathbf{F}-\mathbf{Y}) \right\|_\mathrm{F}^2+\frac{1}{2\sigma}tr(\mathbf{M}^{\top}\mathbf{M}\\
  &-\mathbf{\Lambda}_1^{\top}\mathbf{\Lambda}_1)-\mathbf{\Lambda}_2^{\top}(\mathbf{F1}_c-\mathbf{1}_n)+\frac{\sigma}{2}\left\|\mathbf{F1}_c-\mathbf{1}_n\right\|_2^2\\  J_2(\mathbf{F})=&\beta\left\|\mathbf{F}\right\|_{\mathrm{F}}^2\\
  \end{split}
  \right.\!\!.
\end{equation}
Therefore, in the $t$-the iteration we may linearize $J_2(\mathbf{F})$ at $\mathbf{F}^{(t)}$ via Taylor approximation, and obtain $\tilde{J}_2(\mathbf{F})=\beta\left[\left\|\mathbf{F}^{(t)}\right\|_\mathrm{F}^2+2tr\left(\mathbf{F}^{(t)\top}(\mathbf{F}-\mathbf{F}^{(t)})\right)\right]$. As a result, the approximated objective function $\tilde{J}(\mathbf{F})$ for updating $\mathbf{F}$ is
\begin{equation}\label{eq7}
\begin{split}
   &\tilde{J}(\mathbf{F})=J_1(\mathbf{F}) - \tilde{J}_2(\mathbf{F})\\
  &\ =tr(\mathbf{F}^{\top}\mathbf{LF})\!+\!\alpha\left\|\mathbf{H}\odot(\mathbf{F}\!-\!\mathbf{Y}) \right\|_\mathrm{F}^2\!+\!\frac{1}{2\sigma}tr(\mathbf{M}^{\top}\mathbf{M}
  \!-\!\mathbf{\Lambda}_1^{\top}\!\mathbf{\Lambda}_1)\\
  &\quad -\mathbf{\Lambda}_2^{\top}(\mathbf{F1}_c-\mathbf{1}_n)+\frac{\sigma}{2}\left\|\mathbf{F1}_c-\mathbf{1}_n\right\|_2^2
  -\beta\left[\left\|\mathbf{F}^{(t)}\right\|_\mathrm{F}^2\right.\\
  &\quad \left.+2tr\left(\mathbf{F}^{(t)\top}(\mathbf{F}-\mathbf{F}^{(t)})\right)\right]\\
\end{split}.
\end{equation}

In this paper, we employ the well-known Gradient Descent (GD) method to find the optimal $\mathbf{F}$ that minimizes Eq.~\eqref{eq7}, in which the gradient of $\tilde{J}(\mathbf{F})$ w.r.t $\mathbf{F}$ is computed as
\begin{equation}\label{eq8}
\begin{split}
  \nabla \tilde{J}(\mathbf{F})
  =&2\mathbf{LF}+2\alpha[\mathbf{H}\odot(\mathbf{F}-\mathbf{Y})]-\mathbf{M}-\mathbf{\Lambda}_2\cdot\mathbf{1}_c^{\top}\\
  &\qquad \qquad \qquad \quad ~~ +\sigma(\mathbf{F1}_c-\mathbf{1}_n)\cdot\mathbf{1}_c^{\top}-2\beta\mathbf{F}^{(t)}
\end{split},
\end{equation}
and the updating rule for GD is subsequently $\mathbf{F}:=\mathbf{F}-\tau\nabla \tilde{J}(\mathbf{F})$ with $\tau$ being the stepsize. The detailed CCCP for updating $\mathbf{F}$ in each loop is provided in Algorithm~\ref{alg1}, and the entire ALM optimization process for finding Eq.~\eqref{eq2}'s solution $\mathbf{F}^{\star}$ is summarized in Algorithm~\ref{alg2}. It can be easily verified that the objective function and constraints in Eq.~\eqref{eq2} are twice continuously differentiable, therefore according to \cite{fernandez2012local} the convergence of the ALM process is theoretically guaranteed. Based on $\mathbf{F}^{\star}$, every training example $\mathbf{x}_i$ ($i=1,2,\cdots,n$) will receive its unique valid label as $y_i=\mathop{\arg\max}_{j=1,2,\cdots,c} \mathbf{F}^{\star}_{ij}$, and the corresponding disambiguated label vector is $\overline{\mathbf{F}}_i$ with $\overline{\mathbf{F}}_{ij}=1$ if $j=y_i$ and 0 otherwise.

\begin{algorithm}[t]
   \caption{CCCP for minimizing Eq.~\eqref{eq3}}
   \label{alg1}
\begin{algorithmic}[1]
   \STATE {\bfseries Input:} initial $\mathbf{F}^{(0)}$; stopping criteria $t_{max}=20$, $\epsilon_0=10^{-6}$
   \STATE Set $t=0$;
   \REPEAT
   \STATE Minimizing $\tilde{J}(\mathbf{F})$ in Eq.~\eqref{eq7} via GD;
   \STATE $t:=t+1$;
   \UNTIL $t=t_{max}$ or $\left\|\mathbf{F}^{(t)}-\mathbf{F}^{(t-1)}\right\|_{\mathrm{F}}\leq\epsilon_0$
   \STATE {\bfseries Output:} $\mathbf{F}$ that minimizes Eq.~\eqref{eq3}
\end{algorithmic}
\end{algorithm}

\begin{algorithm}[t]
   \caption{ALM for optimizing Eq.~\eqref{eq2}}
   \label{alg2}
\begin{algorithmic}[1]
   \STATE {\bfseries Input:} training examples $\mathcal{X}\!=\!\{\mathbf{x}_1,\cdots,\mathbf{x}_n\}$ with ambiguous label sets $\mathcal{S}_1,\cdots,\mathcal{S}_n$; tuning parameters $\alpha$, $\beta$, $K$, $\theta$; stopping criteria $loop_{max}=40$, $\epsilon_1=10^{-4}$
   \STATE Construct $K$NN graph $\mathcal{G}$, compute the graph Laplacian matrix $\mathbf{L}$;
   \STATE Compute $\mathbf{Y}$ via Eq.~\eqref{eq1};
   \STATE Set $loop=0$;
   \REPEAT
    \STATE Update $\mathbf{F}$ via CCCP in Algorithm~\ref{alg1};
    \STATE Update $\mathbf{\Lambda}_1$ via Eq.~\eqref{eq4};
    \STATE Update $\mathbf{\Lambda}_2$ via Eq.~\eqref{eq5};
    \STATE Update $\sigma$ via Eq.~\eqref{eq6};
    \STATE $loop:=loop+1$;
   \UNTIL $loop=loop_{max}$ or $\left\|\mathbf{F}^{(loop)}-\mathbf{F}^{(loop-1)}\right\|_{\mathrm{F}}\leq\epsilon_1$
   \STATE {\bfseries Output:} optimal $\mathbf{F}^{\star}$ that minimizes Eq.~\eqref{eq2}
\end{algorithmic}
\end{algorithm}

\subsection{Test Stage}
\label{sec:test}
Given the disambiguated labels $\overline{\mathbf{F}}_1,\overline{\mathbf{F}}_2,\cdots,\overline{\mathbf{F}}_n$ of the $n$ training examples, we predict the label $y_t$ of a test example $\mathbf{x}_t$ via two steps. Firstly, we find the $\mathbf{x}_t$'s $K$ nearest training examples $\left\{\mathbf{x}_{k_i}\right\}_{i=1}^{K}$ in the Euclidean space, and compute the similarity between $\mathbf{x}_t$ and $\left\{\mathbf{x}_{k_i}\right\}_{i=1}^{K}$ (\emph{i.e.} $\left\{\mathbf{W}_{tk_i}\right\}_{i=1}^{K}$) via Eq.~\eqref{eq_Gaussian}. The disambiguated labels of these $K$ training examples are denoted by $\left\{\overline{\mathbf{F}}_{k_i}\right\}_{i=1}^{K}$. Secondly, a soft label vector $\mathbf{F}_t$ is calculated as the weighted sum of $\left\{\overline{\mathbf{F}}_{k_i}\right\}_{i=1}^{K}$ by $\mathbf{F}_t=\sum\nolimits_{i=1}^{K}\mathbf{W}_{tk_i}\mathbf{\overline{F}}_{k_i}$, and $\mathbf{x}_t$'s label is finally decided as $y_t=\mathop{\arg\max}_{j=1,2,\cdots,c} \mathbf{F}_{tj}$.

\section{Experimental Results}
\label{sec:experiments}

\begin{table*}[t]
  \caption{Summary of the adopted datasets.}\label{tab:Datasets}
  \centering
  \begin{tabular}{lccccc}
  \toprule[2pt]
               & $\#$ Examples & $\#$ Features & $\#$ Classes & Average $\#$ labels & Application\\
  \midrule[1pt]
  \emph{Lost}         & 1122      & 512       & 16       & 2.23 & Character-name association in TV Serial\\
  \emph{MSRCv2}       & 591       & 512       & 23       & 1.71 & Ambiguous image classification\\
  \emph{Soccer Player}& 17472     & 279       & 171      & 2.09 & Automatic face naming in news images\\
  \emph{Bird Song}    & 4998      & 38        & 13       & 2.18 & Bird sound classification\\
  \bottomrule[2pt]
\end{tabular}
\end{table*}

In this section, we compare the performances of our proposed RegISL with several existing typical SLL methods on various practical applications such as character-name association in TV show, ambiguous image classification, automatic face naming in news images, and bird sound classification. \par

A variety of methods belonging to different threads mentioned in the introduction (Section~\ref{sec:related_work}) serve as baselines for our comparison, which include
\begin{enumerate}
  \item Regularization-based methods: the compared approaches include SVM-like methodologies MaxiMum Margin Superset Label learning (M3SL) \cite{yuACML15}, Convex Loss for Superset Labels (CLSL) \cite{cour2011learning} and its simplified version with the naive loss (CLSL\_Naive) proposed in \cite{jin2002learning}; Coding theory based methodology Error-Correcting Output Codes (ECOC) \cite{zhang2014disambiguation}; and probability based Logistic Stick-Breaking Conditional Multinomial Model (LSB-CMM) \cite{liu2012conditional}. Note that another typical SVM-like method \cite{nguyen2008classification} is not compared because its accuracy is consistently lower than the latest M3SL with a considerable margin as reported in \cite{yuACML15}.
  \item Instance-based methods: the compared approaches include the traditional Superset Label KNN (SLKNN) \cite{hullermeier2006learning}, and the state-of-the-art method Instance-based Superset Label learning (ISL) \cite{zhangsolving2015}.
\end{enumerate}
For fair comparison, all the above baselines except SLKNN are implemented by using the codes directly provided by the authors. Although the code of SLKNN is not publicly available, it is very easy to reproduce and we implement this algorithm by ourselves.\par

In each of the experiments below, we randomly split the dataset into five non-overlapped folds, and conduct the five-fold cross validation on all the compared methods. In each partition, 80\% examples with their ambiguous labels constitute the training set, and the rest 20\% examples are used for testing. Note that in each partition we keep the ratio of the number of examples from each class in the training set approximately identical to that in the test set. The partitions are also kept identical for all the compared methods. The mean training accuracy and test accuracy averaged over the five different partitions are calculated to assess the classification ability of all the competing algorithms. Besides, we also use the Friedman test \cite{Milton1937The} with 90\% confidence level to investigate whether the proposed RegISL is significantly superior/inferior to the adopted baselines.

\subsection{Character-name Association in TV Serial}
\label{sec:TVSerial}
As mentioned in Section~\ref{sec:intro}, it is meaningful to study how to build the one-to-one correspondence between each character appeared in the video and the real name indicated by the script. To this end, we use the \emph{Lost} dataset provided in \cite{cour2011learning,cour2009learning} to associate the characters in the TV serial ``Lost'' with their groundtruth names. This dataset contains totally 1122 registered face images across 16 characters, and each character has $18\sim204$ images. Given a scene, each of the appeared faces corresponds to an example and it is ambiguously labeled by all the names in the aligned script. The average amount of candidate labels for a single example in this dataset is 2.23. In our experiment, we resize every face image to $30\times 20$ pixels which is further characterized by a 512-dimensional GIST feature \cite{oliva2001modeling}. Please refer to Table~\ref{tab:Datasets} for the details of the adopted datasets.\par

The regularization parameter $C$ in both CLSL and CLSL\_Naive is set to the default optimal value 1000. The maximum value for regularization parameter $C_{max}$ in M3SL is set to 0.01 as recommended by the authors of \cite{yuACML15}. The optimization problems in M3SL are efficiently solved by utilizing the off-the-shelf solvers LIBLINEAR \cite{fan2008liblinear} and CVX \cite{cvx}. In ECOC, the codeword length $L$ is adaptively determined as $L=\lceil100\times log_2(c)\rceil$, where ``$\lceil\cdot\rceil$'' rounds up the inside value to the nearest integer, and $c$ is the number of classes as defined in Section~\ref{sec:intro}. The inherited SVM utilizes the RBF kernel with the width $\gamma=0.5$, and the regularization parameter is $C=5$. In LSB-CMM, the number of mixture components is 10, and the parameter for the involved Dirichlet prior is $\alpha=0.05$ \cite{liu2012conditional}. The balancing parameter $\alpha$ in the iteration expression of ISL is set to 0.9 according to \cite{zhangsolving2015}. For fair comparison, the number of neighbors $K$ in ISL, SLKNN and RegISL is set to the same value 5. In this paper, the two trade-off parameters $\alpha$ and $\beta$ in RegISL are tuned to 1000 and 0.01, respectively. In Section~\ref{sec:ParametricSensitivity}, we will study the sensitivity of these two parameters and also explain why we set $\alpha$ and $\beta$ to these values.\par

The training accuracy and test accuracy obtained by all the algorithms are presented in Table~\ref{tab:Lost}, in which the highest and second highest records are highlighted in red and blue color, respectively. Because SLKNN is a lazy learning algorithm that does not have a training process, its training accuracy is incomputable and thus is not reported. From Table~\ref{tab:Lost} we have some interesting findings:\par

\begin{table}
  \caption{Experimental results on \emph{Lost} dataset. Each record represents ``mean accuracy $\pm$ standard deviation''. The best and second best records are marked in red and blue, respectively. ``$\surd (\times)$'' indicates that RegISL is significantly better (worse) than the corresponding method.}\label{tab:Lost}
  \centering
  \begin{tabular}{lll}
  \toprule[2pt]
             & Training Accuracy & \quad Test Accuracy  \\
  \midrule[1pt]
  CLSL \cite{cour2011learning} & 0.785 $\pm$ 0.016 ${\surd}$ & 0.701 $\pm$ 0.030 ${\surd}$ \\
  CLSL\_Naive \cite{cour2011learning}  & 0.734 $\pm$ 0.017 ${\surd}$ & 0.663 $\pm$ 0.018 ${\surd}$ \\
  ISL \cite{zhangsolving2015}         & \textcolor[rgb]{0.00,0.07,1.00}{0.821 $\pm$ 0.018} ${\surd}$ & \textcolor[rgb]{0.00,0.07,1.00}{0.708 $\pm$ 0.032} \\
  M3SL \cite{yuACML15}         & 0.742 $\pm$ 0.005 ${\surd}$ & 0.668 $\pm$ 0.028 ${\surd}$ \\
  ECOC \cite{zhang2014disambiguation}         & 0.728 $\pm$ 0.013 ${\surd}$ & 0.659 $\pm$ 0.036 ${\surd}$ \\
  LSB-CMM \cite{liu2012conditional}      & 0.782 $\pm$ 0.024 ${\surd}$ & 0.692 $\pm$ 0.021 ${\surd}$     \\
  SLKNN \cite{hullermeier2006learning}        &  \qquad \quad -     & 0.603 $\pm$ 0.020 ${\surd}$ \\
  RegISL       & \textcolor[rgb]{1.00,0.00,0.00}{0.852 $\pm$ 0.011}  & \textcolor[rgb]{1.00,0.00,0.00}{0.726 $\pm$ 0.026} \\
  \bottomrule[2pt]
\end{tabular}
\end{table}

Firstly, the disambiguation operation mentioned in Section~\ref{sec:intro} is critical to improve the performance. We observe that SLKNN and ECOC obtain the lowest test accuracy because they do not contain such disambiguation operation, so the noisy candidate labels of the training examples may impair the training quality and also decrease the test accuracy. CLSL generates higher training accuracy and test accuracy than CLSL\_Naive because CLSL improves CLSL\_Naive by not equally treating all the candidate labels any more. Therefore, CLSL pays more attention to the true positive labels of training examples than CLSL\_Naive during the training stage and produces more satisfactory performance.\par

Secondly, the regularization technique adopted by our RegISL enhances the quality of existing disambiguation operation. Table~\ref{tab:Lost} clearly shows that the proposed RegISL achieves the best performance among all the comparators. The averaged training accuracy and test accuracy are 0.852 and 0.726, respectively. Comparatively, another state-of-the-art instance-based method ISL performs slightly worse than RegISL, which suggests that introducing regularization to instance-based SLL helps to boost the classification accuracy. We think that two factors considered by Eq.~\eqref{eq_addmodel} contribute to the improved performance: one is the smoothness term that models the label similarity between different examples on the graph, and the other one is the discrimination term that highlights the most likely labels from all the possible candidate labels for every training example. These two factors make the entire disambiguation operation of RegISL more accurate than ISL, which further brings about higher training accuracy. Besides, it is straightforward that a better disambiguated training set containing less incorrect labels will lead to more encouraging test performance, that is why our RegISL also obtains the best test accuracy when compared with other baselines.

\subsection{Ambiguous Image Classification}
\label{sec:imageclassification}

\begin{figure}
  \centering
  \includegraphics[width=\linewidth]{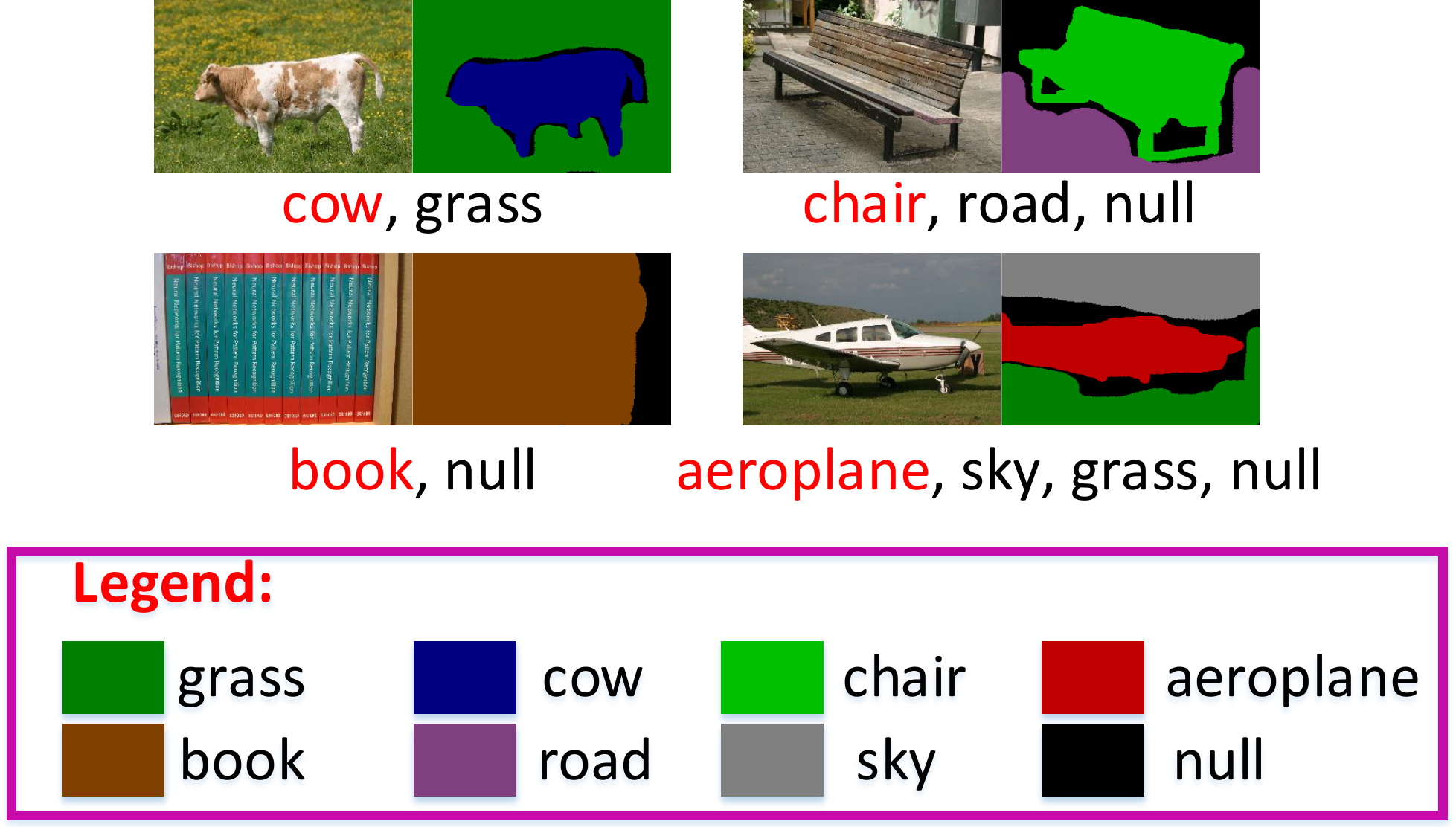}\\
  \caption{Example images of the \emph{MSRCv2} dataset. The labels of segmented regions are regarded as the candidate labels of the entire image (see the labels below the images), in which the label of the most dominant region is determined as the unique groundtruth label of the corresponding image, such as the labels ``cow'', ``chair'', ``book'' and ``aeroplane'' that are marked in red.}\label{fig:MSRCv2Examples}
\end{figure}

\begin{table}
  \caption{Experimental results on \emph{MSRCv2} dataset. Each record represents ``mean accuracy $\pm$ standard deviation''. The best and second best records are marked in red and blue, respectively. ``$\surd (\times)$'' indicates that RegISL is significantly better (worse) than the corresponding method.}\label{tab:MSRC}
  \centering
  \begin{tabular}{lll}
  \toprule[2pt]
             & Training Accuracy & \quad Test Accuracy  \\
  \midrule[1pt]
  CLSL \cite{cour2011learning}      & 0.274 $\pm$ 0.017 ${\surd}$ & 0.208 $\pm$ 0.051 ${\surd}$  \\
  CLSL\_Naive \cite{cour2011learning}& 0.229 $\pm$ 0.019 ${\surd}$ & 0.168 $\pm$ 0.047 ${\surd}$  \\
  ISL \cite{zhangsolving2015}       & \textcolor[rgb]{0.00,0.07,1.00}{0.634 $\pm$ 0.015} ${\surd}$ & \textcolor[rgb]{0.00,0.07,1.00}{0.328 $\pm$ 0.043}  \\
  M3SL \cite{yuACML15}      & 0.398 $\pm$ 0.020 ${\surd}$ & 0.285 $\pm$ 0.025 ${\surd}$   \\
  ECOC \cite{zhang2014disambiguation}      & 0.555 $\pm$ 0.030 ${\surd}$ & 0.251 $\pm$ 0.032 ${\surd}$ \\
  LSB-CMM \cite{liu2012conditional}   & 0.369 $\pm$ 0.007 ${\surd}$ & 0.292 $\pm$ 0.027 ${\surd}$  \\
  SLKNN \cite{hullermeier2006learning}     &  \qquad \quad -    & 0.236 $\pm$ 0.042 ${\surd}$ \\
  RegISL     & \textcolor[rgb]{1.00,0.00,0.00}{0.697 $\pm$ 0.019} & \textcolor[rgb]{1.00,0.00,0.00}{0.333 $\pm$ 0.032}  \\
  \bottomrule[2pt]
\end{tabular}
\end{table}

To test the classification ability of different methods on ambiguous image classification, we follow \cite{liu2012conditional} and \cite{zhang2014disambiguation} and use the \emph{MSRCv2} dataset for our comparison. This dataset contains 591 natural images with totally 23 classes. Every image is segmented into several compact regions with specific semantic information, and the labels of segmented regions form the candidate label set for the entire image. Among the segmented regions, the label of the most dominant region is taken as the single groundtruth label for the given image (see Fig.~\ref{fig:MSRCv2Examples}). Similar to the experiment on \emph{Lost} dataset, we also adopt the 512-dimensional GIST feature to represent the images, and all feature vectors are normalized to unit length for all the competing methodologies.\par

The parameter settings of CLSL, CLSL\_Naive, M3SL, ECOC, and LSB-CMM on \emph{MSRCv2} are the same with those on \emph{Lost} dataset, because they are directly suggested by the authors. The graph parameters $K$ and $\theta$ for ISL, SLKNN and RegISL are respectively set to 10 and 0.1, where the optimal $K$ is chosen from the set $\{5,10,15,20\}$, and $\theta$ is selected from $\{0.01, 0.1, 1, 10\}$. \par

The experimental results are reported in Table~\ref{tab:MSRC}, which reveals that all the methods obtain relatively low accuracy. This is because \emph{MSRCv2} dataset is quite challenging for SLL. Firstly, this dataset is not large, but contains as many as 23 classes (see Table~\ref{tab:Datasets}), so the training examples belonging to every class are very sparse. Besides, the number of examples having a certain candidate label ranges from 24 to 184, therefore such insufficient and skewed training examples pose a great difficulty for training a reliable classifier. Secondly, Fig.~\ref{fig:MSRCv2Examples} reveals that the images in \emph{MSRCv2} are very complex, and the dominant foreground is often surrounded by the background regions with false positive labels, which will mislead both the training and test stages. Although this dataset is quite challenging, Table~\ref{tab:MSRC} clearly indicates that the proposed RegISL still outperforms other methods with a noticeable margin in terms of either training accuracy or test accuracy. Specifically, it can be observed that RegISL leads the second best method ISL with the margins roughly 0.06 on training accuracy and 0.005 on test accuracy, which again demonstrate the superiority of our regularization strategy to the existing non-regularized instance-based model. In contrast, the training accuracy and test accuracy obtained by the remaining approaches like CLSL, CLSL\_Naive, M3SL, ECOC, LSB-CMM and SLKNN do not exceed 0.6 and 0.3, which are much worse than our RegISL.

\subsection{Automatic Face Naming in News Images}
\label{sec:AutomaticFaceNamingInImages}

\begin{table}
  \caption{Experimental results on \emph{Soccer Player} dataset. Each record represents ``mean accuracy $\pm$ standard deviation''. The highest and second highest records are marked in red and blue, respectively. ``$\surd (\times)$'' indicates that RegISL is significantly better (worse) than the corresponding method.}\label{tab:SoccerPlayer}
  \centering
  \begin{tabular}{lll}
  \toprule[2pt]
               & Training Accuracy &  \quad Test Accuracy  \\
  \midrule[1pt]
  CLSL \cite{cour2011learning}      & 0.654 $\pm$ 0.005 $\surd$ & 0.371 $\pm$ 0.004 $\surd$  \\
  CLSL\_Naive \cite{cour2011learning}& 0.648 $\pm$ 0.003 $\surd$ & 0.366 $\pm$ 0.005 $\surd$  \\
  ISL \cite{zhangsolving2015}       & 0.676 $\pm$ 0.003  & \textcolor[rgb]{0.00,0.07,1.00}{0.538 $\pm$ 0.007} \\
  M3SL \cite{yuACML15}      & 0.648 $\pm$ 0.004 $\surd$ & 0.473 $\pm$ 0.005 $\surd$  \\
  ECOC \cite{zhang2014disambiguation}      & \textcolor[rgb]{1.00,0.00,0.00}{0.681 $\pm$ 0.001}  & \textcolor[rgb]{1.00,0.00,0.00}{0.547 $\pm$ 0.004} $\times$ \\
  LSB-CMM \cite{liu2012conditional}   & 0.672 $\pm$ 0.001 $\surd$ &  0.525 $\pm$ 0.003 $\surd$  \\
  SLKNN \cite{hullermeier2006learning}     &  \qquad \quad -   & 0.501 $\pm$ 0.003 $\surd$ \\
  RegISL     & \textcolor[rgb]{0.00,0.07,1.00}{0.678 $\pm$ 0.002} & \textcolor[rgb]{0.00,0.07,1.00}{0.538 $\pm$ 0.001}  \\
  \bottomrule[2pt]
\end{tabular}
\end{table}

It is often the case that in a news collection every image is accompanied by a short textual description to explain the content of this image. Such a news image may contain several faces and the associated description will indicate the names of the people appeared in this image. However, the further information about which face matches which name is not specified. Therefore, in this section we use the \emph{Soccer Player} \cite{zeng2013learning,xiao2015automatic} dataset to test the classification ability of various methods on dealing with news data.\par

The \emph{Soccer Player} dataset is collected by Zeng et al. \cite{zeng2013learning}, which includes the names and images of soccer players from famous European football clubs downloaded from the ``www.zimbio.com'' website. There are totally 8640 images containing 17472 faces across 1579 names. By following \cite{zeng2013learning,xiao2015automatic}, we only retain 170 names that occur at least 20 times, and treat the remaining names as ``Null'' class. As a consequence, the faces appeared in every image are manually annotated using the real names from the descriptions or as ``Null'' class. Each face is represented by a 279-dimensional feature vector describing the 13 interest points (facial landmarks) detected by \cite{sivic2009you}. \par

Table~\ref{tab:SoccerPlayer} reports the experimental results, which reflect that ECOC achieves the best results on this dataset. Regarding the training accuracy, our RegISL is significantly better than CLSL, CLSL\_Naive, M3SL, LSB-CMN, and comparable to ISL and ECOC. For test accuracy, RegISL performs favourably to CLSL, CLSL\_Naive, M3SL, LSB-CMN, and SLKNN. However, it is inferior to the results generated by ECOC. Furthermore, we note that RegISL only falls behind ECOC by 0.003 in training accuracy and 0.009 in test accuracy, and it also generates the top level performance among the compared instance-based methods like SLKNN, ISL and RegISL, so the performance of RegISL is still acceptable on this dataset.

\subsection{Bird Sound Classification}
\label{sec:birdsong}
\begin{table}
  \caption{Experimental results on \emph{Bird Song} dataset. Each record represents ``mean accuracy $\pm$ standard deviation''. The best and second best results are marked in red and blue, respectively. ``$\surd (\times)$'' indicates that RegISL is significantly better (worse) than the corresponding method.}\label{tab:BirdSong}
  \centering
  \begin{tabular}{lll}
  \toprule[2pt]
               & Training Accuracy & \quad Test Accuracy  \\
  \midrule[1pt]
  CLSL \cite{cour2011learning}      & 0.615 $\pm$ 0.003 $\surd$ & 0.414 $\pm$ 0.004 $\surd$  \\
  CLSL\_Naive \cite{cour2011learning} & 0.613 $\pm$ 0.001 $\surd$ & 0.414 $\pm$ 0.003 $\surd$  \\
  ISL \cite{zhangsolving2015}       & \textcolor[rgb]{0.00,0.07,1.00}{0.736 $\pm$ 0.004} $\surd$ & \textcolor[rgb]{0.00,0.07,1.00}{0.559 $\pm$ 0.011} $\surd$   \\
  M3SL \cite{yuACML15}      & 0.658 $\pm$ 0.048 $\surd$ & 0.478 $\pm$ 0.036 $\surd$  \\
  ECOC \cite{zhang2014disambiguation}      & 0.361 $\pm$ 0.013 $\surd$ & 0.359 $\pm$ 0.015 $\surd$ \\
  LSB-CMM \cite{liu2012conditional}   & 0.663 $\pm$ 0.006 $\surd$ & 0.482 $\pm$ 0.022 $\surd$  \\
  SLKNN \cite{hullermeier2006learning}     & \qquad \quad - & 0.552 $\pm$ 0.009 $\surd$ \\
  RegISL     & \textcolor[rgb]{1.00,0.00,0.00}{0.766 $\pm$ 0.008}  & \textcolor[rgb]{1.00,0.00,0.00}{0.583 $\pm$ 0.002}  \\
  \bottomrule[2pt]
\end{tabular}
\end{table}

\begin{figure*}
  \centering
  \includegraphics[width=0.8\linewidth]{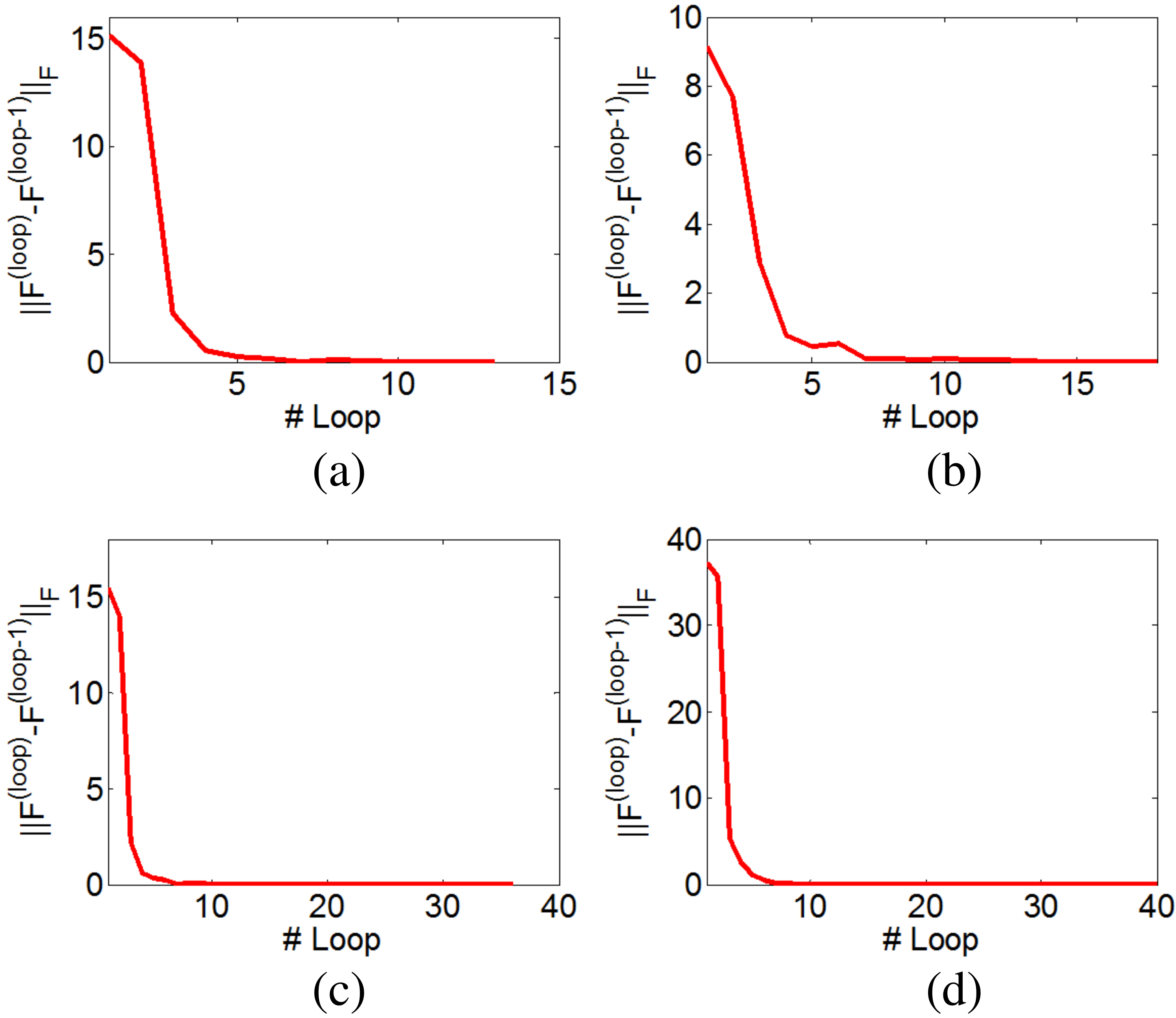}\\
  \caption{The convergence curves of RegISL on the four adopted datasets. (a) is \emph{Lost}, (b) is \emph{MSRCv2}, (c) is \emph{Soccer Player}, and (d) is \emph{Bird Song}.}\label{fig:ConvergenceCurve}
\end{figure*}

In \cite{briggs2012rank}, the authors established a dataset \emph{Bird Song} which contains 548 bird sound recordings that last for ten seconds. Each recording is consisted of 1$\sim$40 syllables, leading to totally 4998 syllables included by the dataset. Each syllable is regarded as an example and is described by a 38-dimensional feature vector. Since every recording contains the songs produced by different species of birds, our target is to identify which example (\emph{i.e.} syllable) corresponds to which kind of bird. In this dataset, the bird species appeared in every record are manually annotated, so they serve as the candidate labels for all the syllables inside this recording.\par

The number of neighbors $K$ for ISL, SLKNN and our RegISL is set to 10, and the kernel width $\theta$ in Eq.~\eqref{eq_Gaussian} is tuned to 1 to achieve the best performance. The trade-off parameters $\alpha$ and $\beta$ are adjusted to 1000 and 0.01 as mentioned in Section~\ref{sec:TVSerial}. We present the training accuracy and test accuracy of all the compared methods in Table~\ref{tab:BirdSong}. A notable fact revealed by Table~\ref{tab:BirdSong} is that the instance-based methods (\emph{e.g.} SLKNN, ISL and RegISL) generate better performance than the regularization-based methodologies such as CLPL, CLPL\_Naive, M3PL, LSB-CMM and ECOC. Among the three instance-based methods, ISL and SLKNN have already achieved very encouraging performances. However, our proposed RegISL can still improve their performances with a noticeable margin regarding either training accuracy or test accuracy. Therefore, the effectiveness of RegISL is demonstrated, which again suggests that integrating the regularization technique with the instance-based framework is beneficial to achieving the improved performance.

\subsection{Illustration of Convergence}
\label{sec:ConvergenceIllustration}

In Section~\ref{sec:model}, we explained that the iteration process of ALM in our algorithm will converge to a stationary point. Here we present the convergence curves of RegISL on the adopted four datasets including \emph{Lost}, \emph{MSRCv2}, \emph{Soccer Player}, and \emph{Bird Song}. In Fig.~\ref{fig:ConvergenceCurve}, the difference of the optimization variable $\mathbf{F}$ between successive loops is particularly evaluated. We observe that the value of $\left\|\mathbf{F}^{(loop)}-\mathbf{F}^{(loop-1)}\right\|_\mathrm{F}$ gradually vanishes when the loops proceed, and the ALM process usually reaches the convergent point between the 13th$\sim$40th loop. Therefore, the convergence of the optimization process employed by our RegISL is demonstrated.

\subsection{Effect of Tuning Parameters}
\label{sec:ParametricSensitivity}

\begin{figure*}
  \centering
  \includegraphics[width=\linewidth]{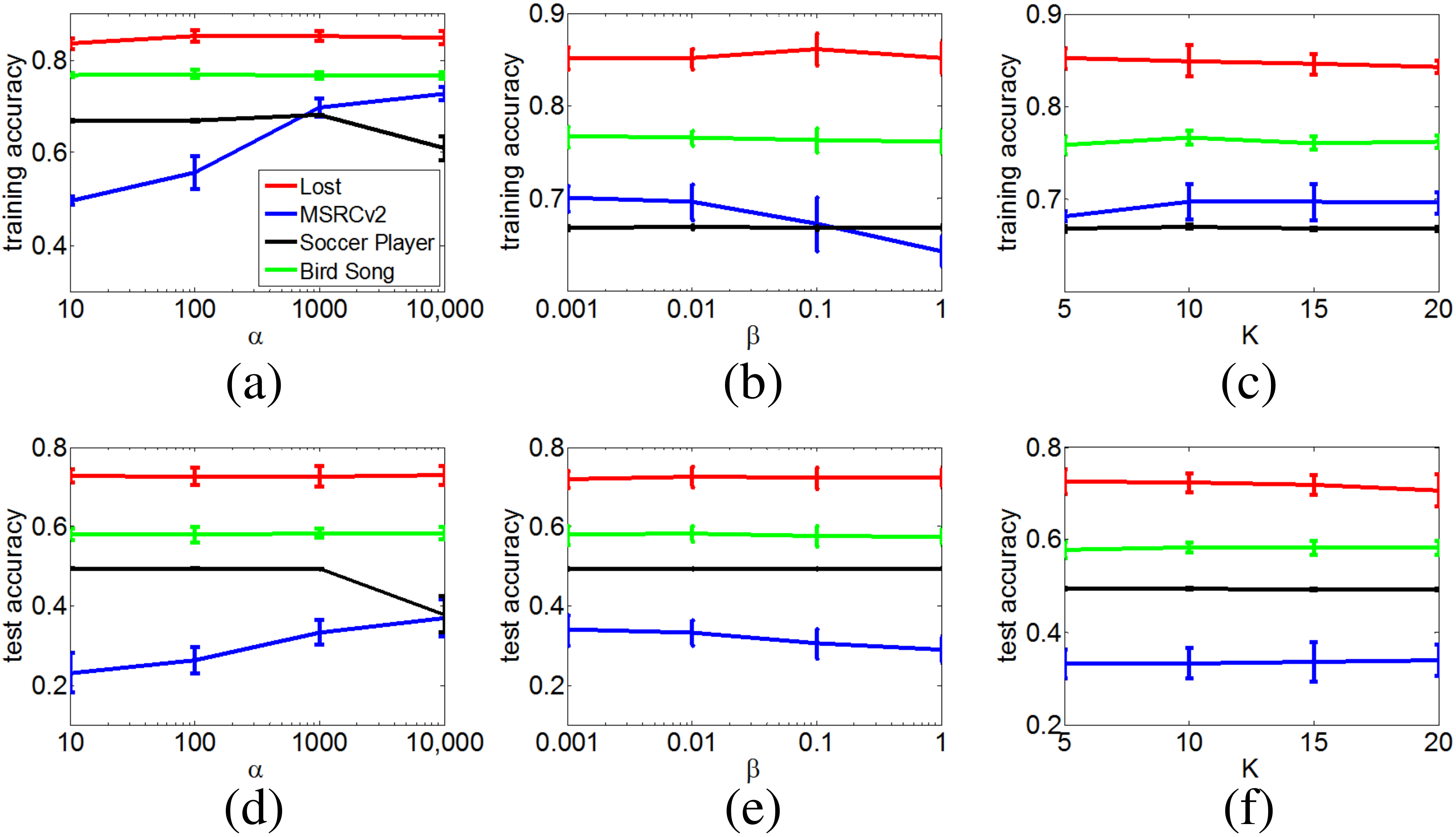}\\
  \caption{Influence of tuning parameters $\alpha$, $\beta$ and $K$ to the final model output on the four datasets. The first column [(a) and (d)] shows the training accuracy and test accuracy obtained by RegISL under different choices of $\alpha$. The second column [(b) and (e)] presents the variations of training accuracy and test accuracy with the increase of $\beta$. The third column [(c) and (f)] plots the training accuracy and test accuracy under different $K$.}\label{fig:ParametricSensitivity}
\end{figure*}

The trade-off parameters $\alpha$, $\beta$ in Eq.~\eqref{eq_addmodel}, and the number of neighbors $K$ are three key tuning parameters in our RegISL model. They should be manually adjusted before implementing the proposed algorithm. Therefore, this section investigates how their variations influence the training accuracy and test accuracy produced by RegISL. The four datasets appeared in Sections~\ref{sec:TVSerial}$\sim$\ref{sec:birdsong} are used here for our experiments.\par

In every dataset, we investigate the effects of $\alpha$, $\beta$ and $K$ on the model output by fixing two of them and then examining the classification accuracy with respect to the change of the remaining one. From the experimental results shown in Fig.~\ref{fig:ParametricSensitivity} we see that the performance of RegISL is generally not sensitive to the choices of these three parameters. In other words, the involved parameters can be easily tuned to achieve satisfactory performance. Specifically, we observe that in most cases RegISL hits the highest accuracy on the four datasets when $\alpha=1000$ and $\beta=0.01$, therefore we use this parameter setting for all the experiments in Sections~\ref{sec:TVSerial}$\sim$\ref{sec:birdsong}. Besides, it can be seen that RegISL obtains the best performance on \emph{Lost}, \emph{MSRCv2}, \emph{Soccer Player} and \emph{Bird Song} datasets when $K=5, 10, 10, 10$, respectively, and this provides us the foundation for choosing the optimal $K$ on the four datasets.

\section{Conclusion}
\label{sec:Conlusion}
In this paper, we propose a novel regularization approach for instance-based superset label learning, which is dubbed as ``RegISL''. Based on the graph $\mathcal{G}$, RegISL disambiguates the candidate labels of training examples by considering both the label smoothness between different examples, and the label discriminative property for every single example. As a consequence, the possible groundtruth labels in the candidate set become manifest while the values of false positive candidate labels are suppressed. 
Thorough experimental results on various practical datasets suggest that in most cases the proposed RegISL achieves better training and test performances than the existing representative SLL methods.\par

Considering that the classification accuracy of our developed RegISL depends on the quality of constructed graph $\mathcal{G}$, in the future we plan to find a way to build a more accurate graph for conducting RegISL. Besides, due to the prevalence of label noise problem \cite{frenay2014classification} today, it would be valuable to extend RegISL to the situation when the groundtruth labels of a small fraction of training examples are not included by their candidate label sets.

%

%
%
%
%
%

\ifCLASSOPTIONcaptionsoff
  \newpage
\fi

{
\bibliographystyle{IEEEtran}
\bibliography{IEEEPartialLabelLearning}
}





\end{document}